\documentclass{article} 
\usepackage{iclr2023_conference,times}

\usepackage{amsmath,amsfonts,bm}

\def\eqref#1{equation~\ref{#1}}

\def\1{\bm{1}}

\DeclareMathAlphabet{\mathsfit}{\encodingdefault}{\sfdefault}{m}{sl}
\SetMathAlphabet{\mathsfit}{bold}{\encodingdefault}{\sfdefault}{bx}{n}

\usepackage{hyperref}
\usepackage{url}
\usepackage{graphicx}
\usepackage{booktabs}
\usepackage{xcolor}
\usepackage{multirow}

\title{Leveraging SAM for Single-Source Domain Generalization in Medical Image Segmentation}

\author{Hanhui Wang, Huaize Ye, Yi Xia, Xueyan Zhang \\
Viterbi School of Engineering\\
University of Southern California\\
Los Angeles, CA 90089, USA \\
\texttt{\{hanhuiwa,huaizeye,xiayi,xueyanz\}@usc.edu} \\
}

\iclrfinalcopy 
\begin{document}
\maketitle

\begin{abstract}
Domain Generalization (DG) aims to reduce domain shifts between domains to achieve promising performance on the unseen target domain, which has been widely practiced in medical image segmentation. Single-source domain generalization (SDG) is the most challenging setting that trains on only one source domain. Although existing methods have made considerable progress on SDG of medical image segmentation, the performances are still far from the applicable standards when faced with a relatively large domain shift. In this paper, we leverage the Segment Anything Model (SAM) to SDG to greatly improve the ability of generalization. Specifically, we introduce a parallel framework, the source images are sent into the SAM module and normal segmentation module respectively. To reduce the calculation resources, we apply a merging strategy before sending images to the SAM module. We extract the bounding boxes from the segmentation module and send the refined version as prompts to the SAM module. We evaluate our model on a classic DG dataset and achieve competitive results compared to other state-of-the-art DG methods. Furthermore, We conducted a series of ablation experiments to prove the effectiveness of the proposed method. The code is publicly available at \url{https://github.com/SARIHUST/SAMMed}.
\end{abstract}

\section{Introduction}

Medical image segmentation plays a crucial role in various healthcare applications, enabling accurate delineation and analysis of anatomical structures or abnormalities within medical images. However, achieving robust and accurate segmentation in medical imaging poses unique challenges due to the complexity and variability of anatomical structures, imaging modalities, noise, and artifacts. In the field of medical image segmentation, domain generalization becomes particularly important due to the significant variations in image styles and characteristics across different medical equipment. Different manufacturers, models, and settings of medical imaging devices can result in distinct image styles even for the same type of target. This inherent diversity makes it essential to develop segmentation algorithms that can effectively generalize across different medical domains, ensuring reliable and accurate segmentation results across a wide range of imaging scenarios.

Since its introduction, the Segment Anything Model (SAM) has shown outstanding performance across a wide spectrum of image segmentation tasks. Thanks to its remarkable flexibility in segmenting objects using diverse prompts, including points, bounding boxes, and text description, SAM possesses the extraordinary capability to perform segmentation tasks across a wide range of scenarios without requiring the burden of additional annotation and training, which makes it well suited for domain generalization tasks. 

However, in fields where precision is of utmost importance, such as the challenging field of Medical Image Processing, it’s worth noting that SAM may not provide the most satisfactory outcomes. Furthermore, SAM relies heavily on the usage of high-quality bounding box prompts to achieve accurate and reliable segmentation results. Unlike regular images, in the context of medical images, the incorporation of such bounding boxes requires a significant level of expertise and domain-specific knowledge. Therefore, while SAM exhibits impressive capabilities in various scenarios, its effectiveness in medical image processing tasks should be evaluated with careful consideration, recognizing the need for further improvements to meet the specific requirements and challenges of the domain.

In contrast, conventional approaches to medical image segmentation do not necessitate the provision of bounding boxes; rather, they heavily rely on meticulously crafted network architectures, which inherently limits their applicability in tasks involving domain generalization. This limitation arises due to the intricate design and manual engineering of these methods, which restricts their ability to adapt and generalize across diverse medical imaging datasets and scenarios.

In order to harness the strengths of both SAM and specific domain knowledge, we now propose a new fine-tuning paradigm. To address the bounding box provision problem of SAM, our approach utilizes a traditional segmentation network to predict coarse masks and then generate refined bounding boxes. We then fine-tune SAM's lightweight mask decoder to generate final prediction masks.

Our main contributions are as follows:
\textbf{(1)} We propose a novel fine-tuning paradigm to leverage SAM for single-sourced domain generalization tasks in medical image segmentation. \textbf{(2)} To fully leverage SAM's image encoder and prompt encoder, we employed image merging and mask filtering techniques to reduce the inference time and enhance overall mask predicting performance. \textbf{(3)} Our approach achieves state-of-the-art performance on the Prostate dataset.

\section{Related Works}

\subsection{Segment Anything Model}
The Segment Anything Model (SAM) proposed by \cite{kirillov2023segment} stands out as a groundbreaking foundational model that introduces the concept of promptability, enabling robust generalization across numerous domains. Trained on the large SA-1B dataset, SAM has demonstrated exceptional performance in zero-shot segmentation of natural images, showcasing its versatility and adaptability. However, when it comes to medical image segmentation, SAM's original model encounters limitations due to its lack of domain-specific knowledge in the realm of medical imaging as demonstrated in \cite{zhang2023segment}.

To address this challenge, previous works have explored various strategies to enhance SAM's performance in medical image segmentation tasks. One such approach involves fine-tuning SAM using medical images, as demonstrated in projects like MedSAM\citep{ma2023segment} and MSA\citep{wu2023medical}. Some of the work like Skinsam\citep{hu2023skinsam} and SAMUS\citep{lin2023samus} applied SAM to more specific segmentation tasks including ultrasound, 3d images, or skin cancer. They aim to impart SAM with a better understanding of medical images' unique characteristics and intricacies, enabling it to produce more accurate and reliable segmentation results. Another promising solution is the utilization of auto-prompting techniques, which have shown promising improvements in  SAM's performance across different domains in projects like  AutoSAM\citep{shaharabany2023autosam} and AutoSAM Adapter\citep{li2023auto}.

Despite these advancements, it is important to note that the current performance achieved by these approaches may not be sufficient for clinical applications that demand higher levels of accuracy and reliability. Additionally, it is worth highlighting that there is no noticeable research focused on the broader aspect of generalization across domains in the context of SAM. While SAM has demonstrated impressive generalization capabilities in various scenarios, there remains a need to explore and develop techniques that enable SAM to effectively generalize across diverse domains, including medical imaging. By addressing this research gap, we can unlock SAM's full potential in tackling the challenges associated with domain generalization and further elevate its applicability in real-world medical image segmentation tasks.

\subsection{Domain Generalization \& Single-Source Domain Generalization}
Image segmentation is critical for computer-aided diagnosis and treatment planning. However, the effectiveness of deep network models trained for specific domains can be hindered when confronted with new, unseen target domains due to unpredictable domain shifts. This challenge has spurred significant interest in the field of domain generalization (DG), which aims to develop models that can generalize well across diverse domains, even those that have not been encountered during training. Several recent works have achieved good performance in medical image segmentation, by applying data augmentation and learning domain-invariant features. So far, some works have applied domain generalization to medical image segmentation. For instance, 
\cite{liu2021feddg}
proposed a method based on meta-learning by encouraging the shape compactness and shape smoothness for prostate segmentation. 
\cite{wang2020dofe} developed a Domain Knowledge Pool to learn and memorize the prior information extracted from multi-source domains, which helps to augment the original image.
\cite{hu2022domain} designed a domain adaptive convolution module and content adaptive convolution module and incorporated both into an encoder-decoder backbone. \cite{zhou2022generalizable} introduced a domain-specific restoration module for regularization and a Random Amplitude Mixup module with low-level frequency knowledge.

Among the various scenarios in DG, single-source domain generalization presents a distinct challenge. Unlike multi-source domain generalization, which benefits from multiple source domains, single-source DG aims to train a predictive model that is both robust and generalizable using only a single source domain. This poses an extreme lack of data, making it more difficult to achieve reliable and accurate segmentation results. Recent research efforts have primarily focused on addressing this challenge through innovative techniques such as data augmentation and the integration of multi-source DG techniques like adversarial training and contrastive learning.\cite{ouyang2022causality} proposed a simple causality-inspired data augmentation approach to expose a segmentation model to synthesized domain-shifted training examples. \cite{su2023rethinking} proposed a location-scale augmentation strategy, that engages the inherent class-level information and enjoys a general form of augmentation. \cite{xu2022adversarial} proposed a mutual information regularizer to enforce the semantic consistency between images from the synthetic domains and estimated the consistency by patch-level contrastive learning. However, the outcomes obtained thus far have been considerably unsatisfying.

\begin{figure}
    \centering
    \setlength{\abovecaptionskip}{-0.5cm}
    \includegraphics[scale=0.4]{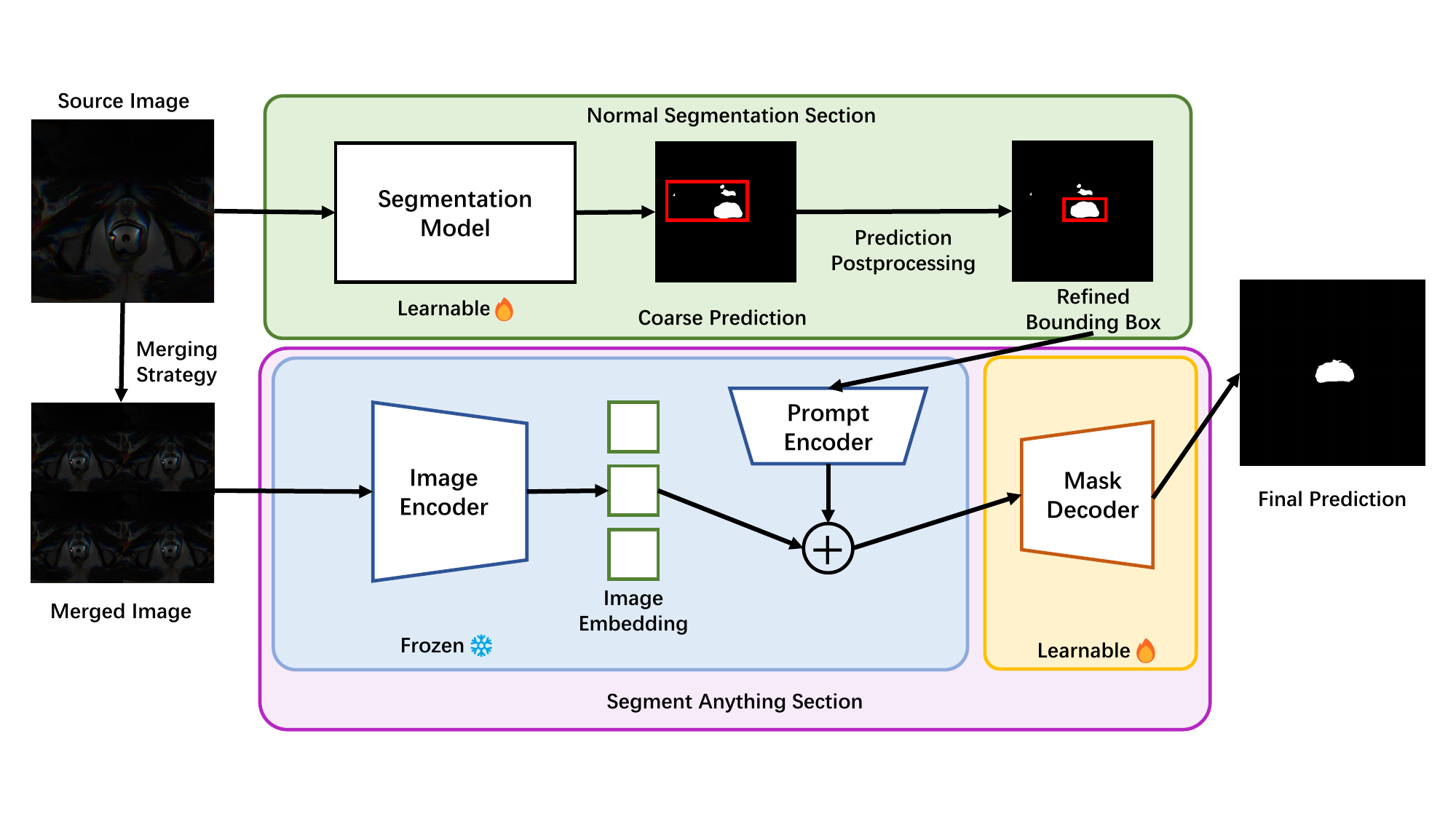}
    \caption{Overall pipeline of our proposed paradigm. Firstly, we utilize a traditional segmentation model to predict coarse masks, and then apply a mask-filtering module to produce refined bounding boxes, which are then used to fine-tune SAM.}
    \label{fig:1}
\end{figure}

\section{Methods}

\subsection{Architecture Overview}
The overview of our methods architecture is shown in Figure \ref{fig:1}. Our approach compromises two primary stages. In the first stage, we utilize a traditional segmentation model to generate coarse prediction masks. We then apply a mask-filtering module to produce refined bounding boxes, which are used in the second stage to fine-tune SAM. Each stage and its main components are described in detail in the following sections.
\subsection{Segmentation Backbone}
\label{sec3.2}
In the first stage, we train a Resnet backbone as a segmentation network to produce coarse prediction masks. The segmentation network is given input of $(I, M_{GT})$ and outputs $M_C$, where $I, M_{GT}$ and $M_C$ denote the source image, ground truth masks, and coarse predictions respectively. Resnet was selected as the backbone network due to its widespread adoption and established reputation in the domain of medical image segmentation.

During the training phase, the network is exclusively trained using data from a single domain. In the subsequent testing phase, the trained network is directly employed on unfamiliar target domains without any additional modifications.

\subsection{Mask-Filtering Module}
\label{sec3.3}
To ensure the production of high-quality bounding boxes, we acknowledge the presence of considerable noise within the coarse masks initially predicted by Resnet. This noise has the potential to significantly degrade the quality of the resulting bounding boxes, which would lead to unpromising mask prediction results from SAM. 

To address this issue, we introduce a mask-filtering module that effectively eliminates noise and retains the largest continuous mask, thereby improving the overall quality of the generated bounding boxes. The underlying concept of this mask-filtering module is straightforward yet impactful. Despite the presence of considerable noise in the output of the segmentation network, it is highly probable that the network accurately identifies the largest region corresponding to the target object. Leveraging this insight, we employ a breadth-first search (BFS) algorithm to identify the largest connected component of the mask, discarding all other components. This refinement process yields improved predictions and facilitates the generation of more precise bounding boxes.

While implementing this module, we observed that the computational speed of the algorithm is significantly influenced by the size of the input image. Considering that the main objective of this step is to eliminate smaller components in the coarse prediction masks, preserving the original size is unnecessary. We devised a strategy to expedite this process by resizing the initial coarse prediction masks $M_C$ to a smaller dimension $M_C'$, (e.g. from $512\times 512$ to $128\times 128$), before executing the algorithm. Subsequently, to restore the original size of the masks, we upscale the corresponding bounding box $B_R$'s parameters by the corresponding factor (e.g. 4 for $512\times 512$ to $128\times 128$). By adopting this approach, we achieve a balance between computational efficiency and the desired accuracy of the bounding boxes.
\subsection{SAM Section}
Following the acquisition of refined bounding boxes, we employed a fine-tuning strategy on SAM, similar to the approach used in MedSAM. During this process, we exclusively fine-tuned SAM's lightweight mask decoder, while keeping the parameters of the image encoder and prompt encoder fixed throughout training. 

However, it is worth noting that MedSAM resizes all images to a standardized dimension of $1024\times 1024$ to accommodate SAM's embedding module. Unfortunately, this resizing operation can incur unnecessary computational overhead and impede training speed in certain situations.

In contrast, to fully exploit the capabilities of SAM's robust image encoder and its multi-box prediction functionality, we adopted a straightforward concatenation method. Specifically, we merged four individual $512\times 512$ images to create a larger composite image with dimensions of $1024\times 1024$. Consequently, the associated bounding boxes $B_R$ and ground truth masks $M_{GT}$ were adjusted accordingly. This merging approach enables us to harness the benefits of SAM's pre-trained embedding modules within its image encoder, while simultaneously expediting the training and testing processes.

Similar to section \ref{sec3.2}, during the training phase, SAM is only fine-tuned using only the data from the same single source domain, while tested on all other target domains.
\section{Experiments}

\subsection{Experiment Setting}
\textbf{Dataset}\ \ \ \ In this study, we assess the performance of our methodology using the publicly available Prostate dataset, which is a diverse collection of T2-weighted MRI prostate images and corresponding masks, spanning six distinct domains.

\textbf{Evaluation Metric}\ \ \ \ We employ a widely adopted evaluation metric in medical image segmentation domains - the Dice coefficient (Dice) metric. The Dice coefficient provides a quantitative measure of the quality of segmentation results, with higher values indicating better segmentation performance.

\textbf{Implementation Details}\ \ \ \ To mitigate the impact of noise in the coarse predictions, we established a higher confidence threshold of $\theta_1=0.75$ for the mask predictions generated by the Resnet backbone. Additionally, considering the superior segmentation performance and increased robustness of the SAM model, we set a lower confidence threshold of $\theta_2=0.5$ for SAM's predictions. In the training process, we use an Adam solver with a base learning rate of 0.0001 and fine-tune SAM for 200 epochs on each domain of the prostate dataset with a batch size of 16.
\subsection{Evaluation on Prostate}
We report the performance of our approach on the Prostate dataset in cross-site evaluation, as listed in Table \ref{tab:1}. Our approach accomplishes the highest Dice score of $79.54\%$, outperforming all previous methods by a large margin. Compared with the former best ACSDG results, our result is $8.12\%$ higher (absolute) and $11.37\%$ better (relative). 

It is also worth noting that although the previous state-of-the-art methods demonstrated impressive performance in certain domains, they exhibited limitations and shortcomings when applied to other domains. In contrast, our method achieves the best performance in all 6 domains, demonstrating a stronger domain generalization ability. 

\begin{table}
    \centering
    \caption{Comparison of Dice Values between our approach and other single-sourced domain generalization methods on the Prostate dataset. – means that the corresponding result was not provided in the paper. The highest values are highlighted in bold fonts, and the second-highest values are marked in red. The data of other methods in the table are collected from CSIDG and ACSDG. The improvement terms indicate our improvement against the previous best results. \textbf{A} stands for training on Domain A and testing on all other domains (same as the \textbf{A to Rest} in later tables).}
    \vspace{1em}
    \begin{tabular}{cccccccc}
        \toprule[1.5pt]
        Method & A & B & C & D & E & F & AVG\\
        \midrule[0.5pt]
        ERM &	71.81 &	65.56 &	43.98 &	71.97 &	48.39 &	37.82 &	56.59\\
        Cutout \cite{devries2017improved}&	78.36&	69.08&	\textcolor{red}{63.45}&	66.39&	61.88&	60.19&	66.56\\
        RSC	\cite{huang2020self}&72.81	&\textcolor{red}{70.18}	&49.18&	\textcolor{red}{74.11}&	54.73&	43.69&	60.78\\
        MixStyle \cite{zhou2021domain}&	73.24&	58.06&	44.75&	66.78&	49.81&	49.73&	57.06\\
        AdvBias \cite{chen2020realistic}&	78.15&	62.24&	54.73&	72.65&	53.14&	51.00&	61.98\\
        RandConv \cite{xu2020robust}&	77.28&	60.77&	53.54&	66.21&	52.12&	36.52&	57.74\\
        CSIDG \cite{ouyang2022causality}&	\textcolor{red}{82.14}&	67.21&	59.11&	73.16&	\textcolor{red}{67.38}&	\textcolor{red}{73.23}&	70.37\\
        ACSDG \cite{xu2022adversarial}&	-&	-&	-&	-&	-&	-&	\textcolor{red}{71.42}\\
        \midrule[0.5pt]
        Ours&	\textbf{84.08}&	\textbf{77.29}&	\textbf{73.98}&	\textbf{82.40}&	\textbf{80.47}&	\textbf{79.04}&	\textbf{79.54}\\
        \midrule[0.5pt]
        Improvements & +1.94 & +7.11 & +10.53 & +8.29 & +13.09 & +5.81 & +8.12\\
        \bottomrule[1.5pt]
    \end{tabular}
    \label{tab:1}
\end{table}

\begin{table}
    \centering
    \caption{Ablation results using different segmentation confidence threshold $\theta_2$ for the SAM predictions. The highest values are highlighted in bold fonts.}
    \vspace{1em}
    \begin{tabular}{cccccccc}
        \toprule[1.5pt]
        Method & A to Rest & B to Rest & C to Rest & D to Rest & E to Rest & F to Rest & Average\\
        \midrule[0.5pt]
        $\theta_2=0.5$& \textbf{0.8408}&	\textbf{0.7729}&	\textbf{0.7398}&	\textbf{0.8240}&	\textbf{0.8047}&	0.7904&	\textbf{0.7954}\\
        $\theta_2=0.75$& 0.8303&	0.7681&	0.7288& 0.8159& 0.7906&	\textbf{0.7929}&	0.7878\\
        $\theta_2=0.9$& 0.8100&	0.7577&	0.7006&	0.7975&	0.7786&	0.7877&	0.7720\\
        \bottomrule[1.5pt]
    \end{tabular}
    \label{tab:2}
\end{table}

\subsection{Ablation Studies}
We conducted a series of ablation experiments on the Prostate dataset to analyze the design and parameter selection in our method.

\textbf{Ablation on the confidence threshold of SAM}\ \ \ \ We used different threshold parameters $\theta_2$ to filter the final prediction of SAM. The performance varies as shown in Table \ref{tab:2}. SAM showcases exceptional precision in mask prediction, achieving highly accurate overall prediction results even when the confidence score exceeds only 0.5. However, when a higher confidence threshold is employed, the model might mistakenly eliminate parts that are, in fact, correct predictions. If not specified, the confidence threshold $\theta_2$ is set to 0.5 as the default setting.

\textbf{Ablation on different modules}\ \ \ \ The quantitative findings of this section are presented in Table \ref{tab:3}. Initially, we fine-tuned SAM using ground truth bounding boxes directly obtained from the ground truth masks, representing the upper limit of our method. This approach yielded outstanding results, with an average Dice score surpassing $90\%$. However, when we employed the most straightforward method of using the full image size as the bounding box input, SAM's performance noticeably decreased to approximately $74.41\%$. This decline highlights the significant influence of bounding box accuracy and quality on the overall segmentation performance of SAM.

To address the need for precise bounding boxes, we applied the segmentation backbone network in section \ref{sec3.2} to generate coarse masks and corresponding bounding boxes to fine-tune SAM. However, when we adopted the bounding boxes extracted from the coarse prediction masks directly, we did not observe any improvement in the results. Upon visual examination, we noticed a substantial presence of noise in the prediction masks (e.g. second row of Figure \ref{fig:2}), which most likely contributed to the inaccurate bounding boxes. Therefore, we further employ the mask-filtering module described in section \ref{sec3.3} to filter out the noise components from each mask, enabling the generation of refined bounding boxes. Fine-tuning SAM using these refined bounding boxes resulted in a significant enhancement of segmentation performance, reaching a Dice score of $79.54\%$.
\begin{table}
    \centering
    \caption{Ablation results on different modules. SAM+GT represents the results when we fine-tune SAM with the ground truth bounding boxes, which is the upper bound of our method and highlighted in red. SAM+Full represents the results when we fine-tune SAM with the full image size as bounding boxes. Resnet represents the results of the coarse prediction masks. SAM+Resnet represents the results when we fine-tune SAM with the coarse mask bounding boxes (without mask filtering). The final results are the same with Table \ref{tab:1}. The highest results (upper bound excluded) are highlighted in bold fonts.}
    \vspace{1em}
    \begin{tabular}{cccccccc}
        \toprule[1.5pt]
        Method & A to Rest & B to Rest & C to Rest & D to Rest & E to Rest & F to Rest & AVG\\
        \midrule[0.5pt]
        \textcolor{red}{SAM+GT}&	\textcolor{red}{91.67}&	\textcolor{red}{92.74}&	\textcolor{red}{90.58}&	\textcolor{red}{90.76}&	\textcolor{red}{92.50}&	\textcolor{red}{90.95}&	\textcolor{red}{91.53}\\
        SAM+Full& 77.82&	\textbf{78.19}&	67.46&	71.04&	78.02&	73.94&	74.41\\
        Resnet&	79.53&	73.02&	68.01&	76.04&	71.55&	65.41&	72.26\\
        SAM+Resnet&	81.76&	75.44&	63.65&	77.14&	63.81&	71.98&	72.30\\
        \midrule[0.5pt]
        Final Result& \textbf{84.08}&	77.29&	\textbf{73.98}&	\textbf{82.40}&	\textbf{80.47}&	\textbf{79.04}&	\textbf{79.54}\\
        \bottomrule[1.5pt]      
    \end{tabular}
    \label{tab:3}
\end{table}

\subsection{Qualitative Results}
\begin{figure}
    \centering
    \setlength{\abovecaptionskip}{-0.3cm}
    \includegraphics[scale=0.4]{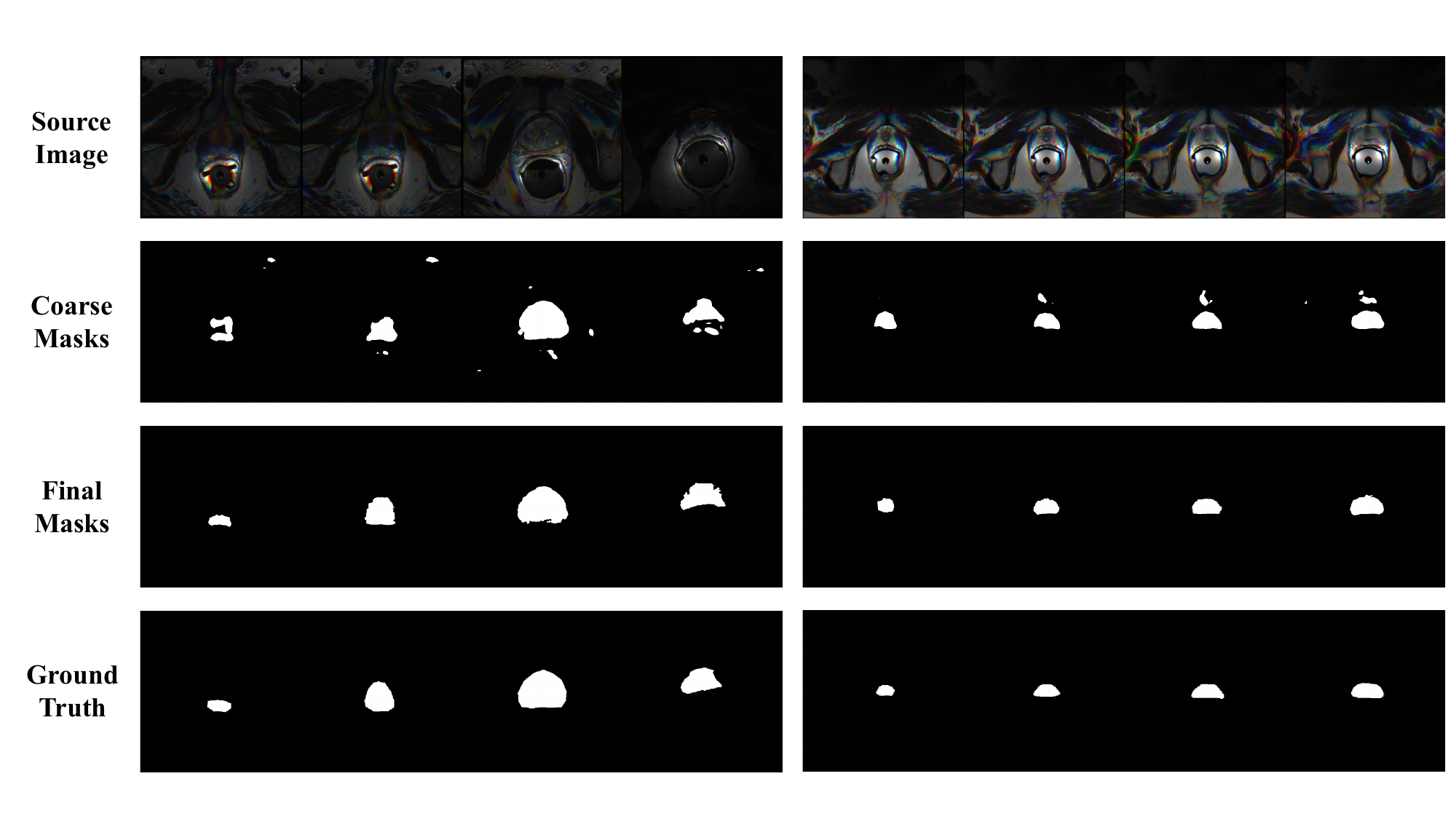}
    \caption{Qualitative results on Domain E and F of the Prostate dataset. The left column demonstrates 4 instances from domain E, and the right column demonstrates 4 instances from domain F. The first row demonstrates the source image for segmentation, the second row demonstrates the coarse masks predicted by the Resnet we use as the segmentation model, the third row demonstrates the final mask results predicted by our fine-tuned SAM, and the last row demonstrates the ground truth masks.}
    \label{fig:2}
\end{figure}

Figure \ref{fig:2} shows the qualitative outcomes of fine-tuning SAM using our approach on Domain E and Domain F of the Prostate dataset. We randomly selected four instances from each domain and applied SAM, fine-tuned on Domain F, to segment instances in Domain E, and vice versa. The results demonstrate that the process of predicting coarse masks and applying our mask-filtering module to generate refined bounding boxes can effectively enhance the accuracy of SAM during the fine-tuning process.

\section{Conclusion}

In conclusion, we have presented a novel framework that integrates SDG and SAM with mask-filtering postprocessing and merging strategy. We demonstrated the effectiveness of all the proposed modules and evaluated our approach to the Prostate dataset. The experiment showed that our method had greatly exceeded the state-of-the-art SDG methods in all directions of generalization on the dataset we trained. For future improvement, we plan to investigate more strategies to further reduce calculation costs and apply the model to other segmentation tasks while enhancing the network's performance.

\bibliography{iclr2023_conference}
\bibliographystyle{iclr2023_conference}


\end{document}